\documentclass[final]{cvpr}

\usepackage{times}
\usepackage{epsfig}
\usepackage{graphicx}
\usepackage{subfigure}

\usepackage{amsmath,amssymb,amsfonts,dsfont,bm,mathrsfs}
\usepackage{booktabs,multirow,adjustbox}
\usepackage[pagebackref=true,breaklinks=true,colorlinks,bookmarks=false]{hyperref}
\makeatletter
\@namedef{ver@everyshi.sty}{}
\makeatother
\usepackage{tikz}
\newcommand\buildcirclesign[2]{%
    \begin{tikzpicture}[baseline=(X.base), inner sep=0, outer sep=0]
    \node[draw,circle] (X)  {\ensuremath{#1 #2}};
    \end{tikzpicture}%
}
\newcommand{\circlesign}[1]{ 
    \mathbin{
        \mathchoice
        {\buildcirclesign{\displaystyle}{#1}}
        {\buildcirclesign{\textstyle}{#1}}
        {\buildcirclesign{\scriptstyle}{#1}}
        {\buildcirclesign{\scriptscriptstyle}{#1}}
    } 
}

\newcommand{\W}{{\rm\bf W}}  
\newcommand{\X}{{\rm\bf X}}  

\def\cvprPaperID{9} 
\def\confYear{CVPR 2021}

\begin{document}

\title{CompConv: A Compact Convolution Module for Efficient Feature Learning}



\author{
  Chen Zhang$^1$ \quad Yinghao Xu$^2$ \quad Yujun Shen$^2$ \\
  $^1$Zhejiang University \quad $^2$The Chinese University of Hong Kong \\
  {\tt\small daisy\_chen@zju.edu.cn \quad \{xy119, sy116\}@ie.cuhk.edu.hk}
}
\maketitle

\begin{abstract}
Convolutional Neural Networks (CNNs) have achieved remarkable success in various computer vision tasks but rely on tremendous computational cost.
To solve this problem, existing approaches either compress well-trained large-scale models or learn lightweight models with carefully designed network structures.
In this work, we make a close study of the convolution operator, which is the basic unit used in CNNs, to reduce its computing load.
In particular, we propose a compact convolution module, called CompConv, to facilitate efficient feature learning.
With the divide-and-conquer strategy, CompConv is able to save a great many computations as well as parameters to produce a certain dimensional feature map.
Furthermore, CompConv discreetly integrates the input features into the outputs to efficiently inherit the input information.
More importantly, the novel CompConv is a plug-and-play module that can be directly applied to modern CNN structures to replace the vanilla convolution layers without further effort.
Extensive experimental results suggest that CompConv can adequately compress the benchmark CNN structures yet barely sacrifice the performance, surpassing other competitors.
\end{abstract}
\section{Introduction}\label{sec:introduction}
In recent years, Convolutional Neural Networks (CNNs) have significantly advanced many tasks in the computer vision field~\cite{resnet, alexnet, vgg} due to their great power in learning representative features.
However, such success relies on massive computation and storage resources, hindering these large-scale models from being deployed on resource-limited devices in practice.

To alleviate the pressure on the demand for computing power and memory, one straight-forward solution is to compress well-trained big models.
For this purpose, existing approaches propose either to employ a tiny model as the student to distill the knowledge from the teacher model~\cite{distilling, co-teaching}, or to prune the unimportant connections from the learned model to make it thinner~\cite{li2016pruning, he2017channel, han16pruning}.
Besides training large-scale models and then performing compression, an alternative solution is to directly learn a lightweight model.
Nevertheless, a side effect of reducing the model size is the sacrifice of the learning capacity.
To tackle this obstacle, it often requires careful design of the network architecture, such as MobileNet~\cite{mobilenets, mobilenetv2} and SqueezeNet~\cite{squeezenet}.

\begin{figure}[t]
    \centering
    \subfigure[Vanilla Convolution]{
    \includegraphics[width=0.45\columnwidth]{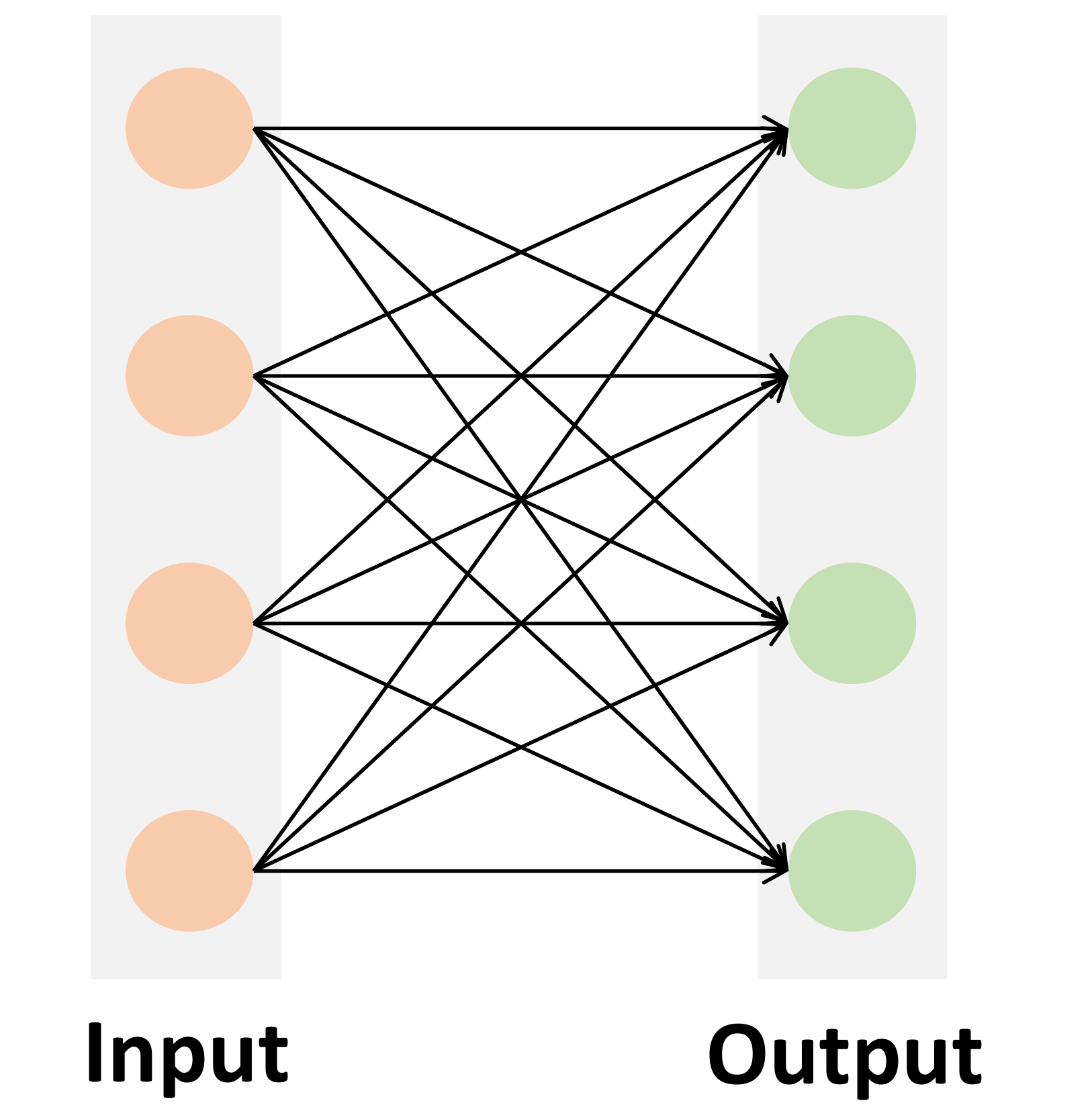}}
    \subfigure[CompConv]{
    \includegraphics[width=0.45\columnwidth]{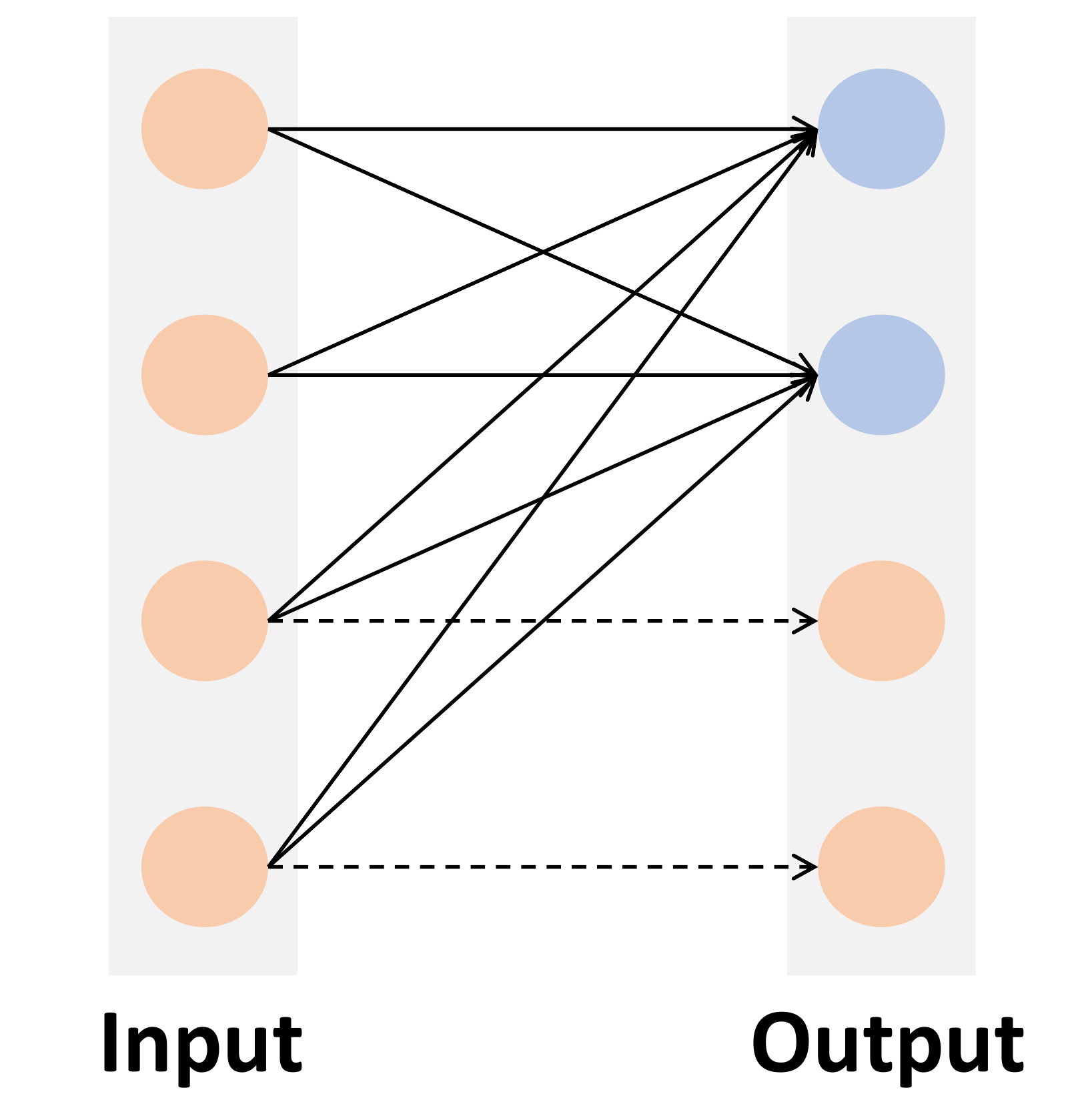}}
    \vspace{0pt}
    \caption{
        Concept diagram of the proposed compact convolution module, CompConv.
        Compared to the vanilla convolution in (a), which fully connects the input channels with the output channels, CompConv adopts the divide-and-conquer strategy to develop the output recursively.
        (b) gives a brief illustration of the core unit used in CompConv, which only uses the convolution operation to contribute half number of channels (\textit{i.e.}, the solid line) and subtly borrows the other half from the inputs (\textit{i.e.}, the dashed line).
    }
    \label{fig:diagram}
\end{figure}

In this paper, different from prior work, we take a close look into the convolution operator, which is the basic unit used in all CNNs, and propose to save the computing load by compressing the convolution module.
Recall that the convolution module learns the transformation from one feature space to another.
More concretely, given an input feature map, with $C_{in}$ channels, the convolutional kernel projects it to a $C_{out}$-channels feature map as the output.
For each pixel in the spatial field, the conventional convolution requires $C_{in} \times C_{out}$ connections, as shown in Fig.~\ref{fig:diagram}a.
As pointed out by prior work~\cite{ghostnet, li2016pruning, han16pruning}, however, the learning of CNNs presents a definite over-parameterization and redundancy.
From this perspective, we propose \textit{CompConv} to implement the convolution more compactly.

Specifically, instead of directly producing the final feature map, CompConv employs a core unit that equally splits the output into two parts along the channel axis.
One of them is transformed from the input feature via convolutional projection while the other identically borrows a subset of input channels.
In this way, we are able to pass down the learned information across the convolution module to the most extent yet with minor effort.
Fig.~\ref{fig:diagram}b illustrates the diagram of such process.
More importantly, our core unit can be performed in a recursive computing manner, resulting in a divide-and-conquer strategy.
As a result, CompConv is able to save a great many computations as well as parameters for more efficient feature extraction.
We summarize our contributions as follows:
\begin{itemize}
    \item We propose a compact convolution module, termed as \textit{CompConv}, which utilizes the divide-and-conquer strategy together with the carefully designed identical mapping to considerably reduce the computational cost of CNNs.
    \item We make exhaustive analysis of the proposed CompConv by studying how the learning capacity is affected by the depth of the recursive computation. We further propose a practical scheme to convincingly control the compression rate.
    \item We apply CompConv to various benchmark CNNs as a handy replacement of the conventional convolution layer. It turns out that CompConv can substantially save the computing load yet barely sacrifice the model performance on classification and detection tasks, outperforming existing approaches.
\end{itemize}

\section{Related Work}\label{sec:related-work}

\vspace{2pt}
\noindent\textbf{Model Compression.}
Modern deep neural networks have millions of weights, rendering them both memory-intensive and computationally expensive.
A straightforward way to reduce the computational cost is to remove the unimportant neuron connections with rule-based or learning-based methods, named network pruning~\cite{han16pruning, li2016pruning, he2017channel, luo2017thinet, Liu2017learning, dong2017learning, luo2017thinet, abcprunner, nisp, sss, HRANK, gal, gdp}.
Recent work also shows that precision computation is not necessary for the training and inference of deep models~\cite{limitednumerical}.
As a result, quantizing weights and activations~\cite{jacob2018quantization,zhou2017incremental} are also widely explored to improve network efficiency.
Specifically, binary networks~\cite{courbariaux2015binaryconnect, rastegari2016xnor}, which only employ 1-bit neuron to represent the model weights and activations, replace all MAC operation with boolean operation to save the computation.
Besides, knowledge distillation~\cite{distilling, co-teaching, zagoruyko2016paying, huang2017like, yim2017gift, yim2017gift} offers an alternative way to generate small student networks with the guidance of a well-trained large teacher network.
In this way, the students show comparable performance as the teacher but raise a more efficient inference process.

\vspace{2pt}
\noindent\textbf{Lightweight Model Design.}
Another way of reducing the computational cost of CNNs is to directly learn a lightweight model, but there is a clear trade-off between the model size and the learning capacity.
As a result, existing approaches propose to increase the model efficiency and maintain its performance at the same time with carefully designed network structure~\cite{squeezenet,mobilenets,mobilenetv2,shufflenet,shufflenetv2,mobilenetv3}.
SqueezeNet~\cite{squeezenet} adopts a large mount of $1\times1$ convolutions to reduce the number of parameters.
MobileNet V1~\cite{mobilenets} and V2~\cite{mobilenetv2} employ depth-wise separable convolutions and inverted linear residual bottleneck to improve computation efficiency.
ShuffleNet V1~\cite{shufflenet} and V2~\cite{shufflenetv2} propose the channel-shuffle operation to enhance the information flow between different channel groups and provide a hardware-friendly implementation to enable practical applications.
GhostNet~\cite{ghostnet} considers the feature redundancy between feature maps and proposes to learn ghost features with cheap operations.
On the other hand, Network Architecture Search (NAS)~\cite{liu2018progressive, zoph2018learning, zoph2016neural,zhong2018practical} aims at finding the most efficient network structure automatically.
MobileNet V3~\cite{mobilenetv3} utilizes the Auto-ML technology~\cite{bender2018understanding,cai2019once} to achieve better performance with fewer floating point operations (FLOPs).
\section{Compact Convolution Module}\label{sec:method}
In this section, we introduce the proposed CompConv module, which efficiently learns the output feature from the input using the divide-and-conquer strategy.
Sec.~\ref{subsec:motivation} introduces the motivation.
Sec.~\ref{subsec:core-unit} describes the core unit used in CompConv.
Sec.~\ref{subsec:recursive-compuation} presents the complete CompConv module, which executes the core unit in a recursive fashion.
Sec.~\ref{subsec:separation-strategy} provides an adaptive strategy of using CompConv in practice.
Sec.~\ref{subsec:complexity-analysis} analyzes the computing complexity of CompConv.

\subsection{Motivation}\label{subsec:motivation}
Convolution can be treated as an operation that maps features from one space to another. 
To some extent, this process is similar to Discrete Fourier Transform (DFT), which maps a signal sequence from the time domain to the frequency domain.
Fast Fourier Transform (FFT) is widely used to speed up the computation of DFT.
Motivated by FFT, our CompConv is proposed to compress the vanilla convolution module through the divide-and-conquer strategy.\footnote{Note that the improvement from DFT to FFT does not change the computational results, \textit{i.e.}, they are identical. Differently, the change from the vanilla convolution to CompConv is not a lossless compression. We simply design CompConv by drawing lessons from FFT for more efficient feature learning.}

Let us review the formulation of FFT.
When applying DFT to a $N$-points signal sequence $x(t)$ from the time domain, FFT proposes to split it into two $\frac{N}{2}$-points sub-sequences, denoted as $x^{(e)}(t)$ and $x^{(o)}(t)$, and perform DFT on each of them.
Here, $e$ and $o$ stand for ``even'' and ``odd'' respectively.
Accordingly, the final result $X(k)$ from the frequency domain can be obtained from the intermediate transformation results $X^{(e)}(k)$ and $X^{(o)}(k)$ with
\begin{align}
    X(k) = X^{(e)}(k) + W^k_N X^{(o)}(k), \label{eq::FFT}
\end{align}
where $W^k_N=\exp(-j\frac{2\pi}{N}k)$ is a multiplier.
Based on this, the factorized results $X^{(e)}(k)$ and $X^{(o)}(k)$ can be further divided into smaller groups, resulting in a recursive computing manner.

\subsection{Core Unit of CompConv}\label{subsec:core-unit}
Inspired by FFT, we introduce the divide-and-conquer strategy into the convolution module used in CNNs to improve its computing efficiency.
By analogy, we treat the intermediate feature maps produced by CNNs as the sequence from the \textit{channel} axis.
More concretely, to develop a feature map $\X$ with $C$ channels, we can alternatively develop two feature maps $\X_A$ and $\X_B$, each of which is with $\frac{C}{2}$ channels, and compose them together with
\begin{align}
    \X = \X_A \circlesign{+} \W \X_B, \label{eq::unit}
\end{align}
where $\circlesign{+}$ denotes the concatenating operation along the channel axis and $\W$ is a learnable parameter used to transform feature maps.

Eq.~\eqref{eq::unit} embodies the key idea of CompConv.
In particular, the core unit of CompConv is implemented with two parts, as shown in Fig.~\ref{fig:conv}.
One part (\textit{i.e.}, $\X_A$) is identically mapped from a subset of input channels, which is able to inherit information from the input with minor effort.
The other part (\textit{i.e.}, $\X_B$) is transformed from the input feature with a native convolution module.

\subsection{Recursive Computation}\label{subsec:recursive-compuation}
With the formulation in Eq.~\eqref{eq::unit}, CompConv can be computed in a recursive manner by further splitting $\X_B$ into two halves as
\begin{align}
    \X_{B_i}= \X_{A_{i+1}} \circlesign{+} \W_{i+1} \X_{B_{i+1}} \quad i = 0,\cdots,d-1, \label{eq::casFFT}
\end{align}
where $d$ denotes the recursion depth.

\vspace{2pt}
\noindent\textbf{Tailing Channels.}
We treat the first separation step (\textit{i.e.}, $\{\X_{A_0}, \X_{B_0}\}$) differently from other steps, as shown in Fig.~\ref{fig:conv}.
Concretely, $\X_{A_0}$ is not directly borrowed from the input but transformed from $\X_{B_0}$ instead.
There are mainly two reasons in doing so.
On one hand, $\X_{A_0}$ is with the most channels among all identical parts $\{\X_{A_i}\}_{i=0}^{d-1}$.
If we directly duplicate some input channels as $\X_{A_0}$, there will be too much redundancy between the input feature map and the output one, severely limiting the learning capacity of this module.
On the other hand, besides transformed from $\X_{B_0}$, there are some alternative ways to obtain $\X_{A_0}$, such as mapping from the entire input feature map or building another recursion like the computation of $\X_{B_0}$.
Among all these approaches, developing $\X_{A_0}$ from $\X_{B_0}$ is with the cheapest computing cost.
Meanwhile, the deduction of $\X_{B_0}$ has already assembled enough information from the input feature, hence the learning capacity can also be maintained.
We use group convolution with group numbers equal to channel number for this transformation.

\vspace{2pt}
\noindent\textbf{Integrating Recursive Results.}
To better utilize the computations in the recursive process, the final output is developed by not only grouping the two largest sub-features (\textit{i.e.}, $\X_{A_0}$ and $\X_{B_0}$) but also integrating all the intermediate results, as shown in Fig.~\ref{fig:conv}.
In this way, we can make sufficient use of all the computing operations to produce the final output. 
In addition, a shuffle block is added after the concatenation of these feature maps.

\definecolor{atomictangerine}{rgb}{1.0, 0.6, 0.4}
\definecolor{darkseagreen}{rgb}{0.56, 0.74, 0.56}
\definecolor{babyblueeyes}{rgb}{0.63, 0.79, 0.95}
\begin{figure}[t]
    \centering
    \includegraphics[width=\columnwidth]{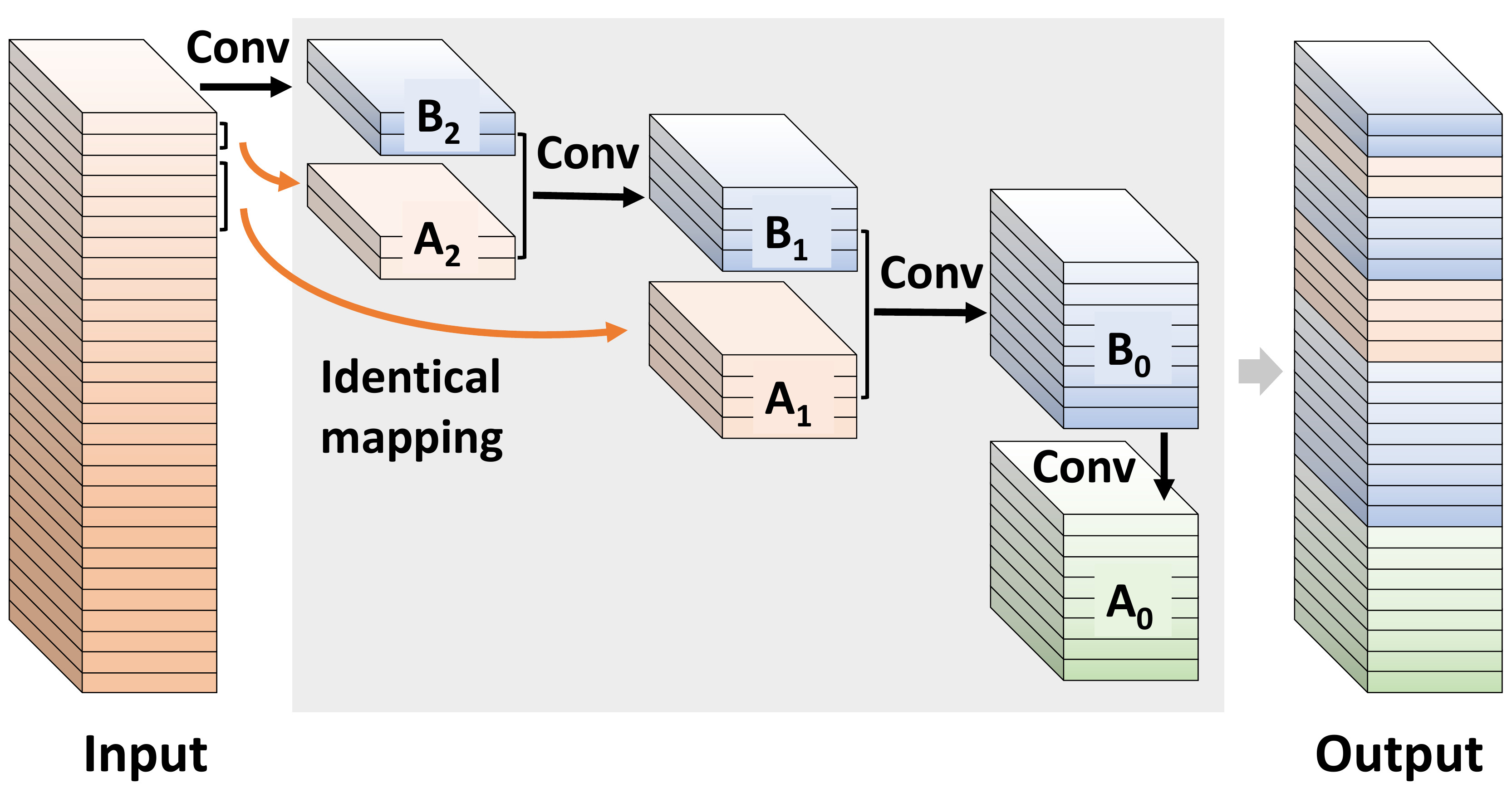}
    \caption{
        Framework of the proposed CompConv with the recursion depth $d=3$.
        With the divide-and-conquer strategy, CompConv computes the output from the input recursively.
        In each step, to get a feature map with a certain number of channels, the core unit of CompConv proposes to generate half of the channels (\textbf{\textcolor{babyblueeyes}{blue}} blocks) with the vanilla convolution operation and the other half (\textbf{\textcolor{atomictangerine}{tangerine}} blocks) with the identical mapping from the input.
        Meanwhile, to increase the learning capability yet maintain low computing load, the identical mapping is replaced with an additional convolution module in the first separation step to produce tailing channels (\textbf{\textcolor{darkseagreen}{green}} blocks).
    }
    \label{fig:conv}
\end{figure}

\subsection{Adaptive Separation Strategy}\label{subsec:separation-strategy}
As described in Sec.~\ref{subsec:recursive-compuation}, CompConv adopts the divide-and-conquer strategy for efficient feature learning.
Thus how to recursively split the channels is critical to its computing efficiency and learning capacity.
Here, we use $C_{in}$ and $C_{out}$ to denote the number of input channels and output channels respectively.
$C_{prim}$ refers to the channel number of the minimum computing unit, \textit{e.g.}, $\X_{B_2}$ when $d=3$ in Fig.~\ref{fig:conv}.
Considering the exponential growth of the channel number during the recursive computation, we can expect
\begin{align}
    C_{out}=\sum^d_{i=1}2^iC_{prim}. \label{eq::channel_sum}
\end{align}

We can easily solve Eq.~\eqref{eq::channel_sum} with
\begin{align}
    C_{prim}=\lceil \frac{C_{out}}{2\times(2^d-1)} \rceil, \label{eq::compute_p}
\end{align}
where $\lceil\cdot\rceil$ denotes the ceiling function that grounds $C_{prim}$ as an integer.
If the sum of the channels from all units is larger than $C_{out}$, we simply drop some last channels in $\X_{A_0}$ to make sure the output feature is with the proper dimension.

\vspace{2pt}
\noindent\textbf{Choice of Recursion Depth.}
We can tell from Eq.~\eqref{eq::compute_p} that $C_{prim}$ is highly dependent of the recursion depth $d$, which is a hyper-parameter in our CompConv module.
Larger $d$ correspond to higher compression rate, where $d=0$ means no compression.
Considering the different structures and different model scales of modern CNNs, we propose an adaptive strategy on choosing the depth as
\begin{align}
    d = \min(\log_2(\max(1, \frac{C_{in}}{C_0})) + 1, 3). \label{eq::recommond_N}
\end{align}
Here, $C_0$ is a model-specific design choice, valuing from $\{32,64,128,256,512,\cdots\}$, which can be determined by the target compression rate and the model size.
Intuitively, larger $C_0$ tends to induce smaller $d$, further resulting in a lighter compression.
From this perspective, $C_0$ can be used to control the trade-off between computing efficiency and learning capacity.
It's worth noting that the recursion depth $d$ is related to the number of input channels, $C_{in}$, in Eq.~\eqref{eq::recommond_N}, which means our strategy will dynamically adjust the computation depth at different layers.
Meanwhile, to guarantee the minimum unit to have adequate learning capacity, we assign it with enough channels.
In other words, $C_{prim}$ could not be too small.
From Eq.~\eqref{eq::compute_p}, we see that when $d=3$, $C_{prim}$ only accounts for $\sim 8\%$ of output channels. 
we therefore bound the depth $d$ with a maximum value $3$.

\vspace{2pt}
\noindent\textbf{Recommended Configuration.}
For most popular CNN networks like VGG~\cite{vgg} and ResNet~\cite{resnet}, we recommend to set $C_0=128$.
We denote this setting as CompConv128.
In the following sections, we use CompConv128 by default if no special instructions.

\subsection{Complexity Analysis}\label{subsec:complexity-analysis}
In this part, we briefly discuss how CompConv can help to save the computing load.
We conduct the complexity analysis and make a comparison between vanilla convolution and the proposed CompConv.
Here, we focus on the number of operations executed in each module.

Assuming both the input and output feature maps are with resolution $H \times W$, the computing complexities for vanilla convolution and CompConv are
\begin{align}
    \mathcal{O}_{Conv} = &\ H \times W \times k^2 \times C_{in} \times C_{out}, \label{eq::complexity-conv} \\
    \mathcal{O}_{CompConv} = &\ H \times W \times k^2 \times (C_{in} \times C_{prim} + \nonumber \\
                             &\sum\limits_{i=1}^{d-1} (2^iC_{prim})^2 + 2^{d-1}C_{prim}), \label{eq::complexity-compconv}
\end{align}
where $k$ indicates the size of the convolutional kernel.
Under the setting $C_{in}=C_{out}$ and $d=3$, CompConv only requires $\sim 20\%$ computational resources compared to the conventional convolution to develop the output feature with the same number of channels.

\section{Experiments}
In this section, we evaluate the effectiveness and efficiency of the proposed CompConv on various visual benchmarks, 
including CIFAR-10~\cite{cifar}, CIFAR-100~\cite{cifar} and ImageNet~\cite{imagenet}. 
Top-1 and Top-5 accuracy are reported as the evaluation metric for comparison. Ablation studies on various CompConv designs are also included.
Besides, detection results on COCO~\cite{coco} are reported to show the efficiency and generality of CompConv.  
The standard mean average precision (mAP) is used to measure accuracy.

This section is organized as follows:
Sec.~\ref{training-setting} introduces the experiment setting, including datasets and the details of training and inference.
Sec.~\ref{CIFAR-10-setting} presents the ablation study on the detailed setting, especially the recursive depth of our CompConv.
Sec.~\ref{main_results} and Sec.~\ref{detection} evaluate the well designed CompConv on various visual benchmarks.
 
\subsection{Experimental Settings} \label{training-setting}
\vspace{2pt}
\noindent{\textbf{Dataset.}} ImageNet~\cite{imagenet} contains around 1.28 million training images and 5k validation images. 
It includes 400 image categories in total. 
CIFAR-10~\cite{cifar} consists of 50k training images and 10k testing images in 10 classes.
The CIFAR-100~\cite{cifar} increases the number of classes to 100, which is a finer version of CIFAR-10~\cite{cifar}.
The COCO benchmark~\cite{coco} contains 118k images for training, 5k images for validation.

\vspace{2pt}
\noindent{\textbf{Training and Inference.}}
For ImageNet classification, we adopt data augmentation scheme including random crop and mirroring following~\cite{resnet}. 
We use the momentum of 0.9 and synchronized SGD training over 8 GPUs.
Each GPU has a batch-size of 32, resulting in a mini-batch of 256 in total.  
The learning rate is 0.1 and will be reduced by a factor of 10 at 30, 60, 90 epochs (100 epochs in total), respectively.
At inference time, the shorter side of samples from ImageNet is resized to 256 and then cropped with 224 $\times$ 224 at the center region. 

For CIFAR-10 and CIFAR-100, we adopt simple data augmentation for training: 
4 pixels are padded on each side, and a 32$\times$32 crop is randomly sampled from the padded image or its horizontal flip. 
We start with a learning rate of 0.1 with a total batch size of 128, divide it by 20 at 60, 120 epochs (150 epochs in total).
And we feed the single view of original images to the network at inference time.

For object detection on COCO~\cite{coco}, the input images are resized to a short side of 800 and a long side without exceeding 1333 following~\cite{mmdetection}.
The models are initialized by the pre-trained ImageNet model and trained on 8 GPUs with 2 images per GPU for 12 epochs (1$\times$ settings). 
In SGD training, the learning rate is initialized to 0.01 and then divided by 10 at epochs 8 and 11. 
The weight decay and momentum parameters are set to $10^{-4}$ and 0.9, respectively.

\subsection{Ablation Study}\label{CIFAR-10-setting}

We choose VGG16~\cite{vgg} as the basic backbone network to conduct the following ablation experiments on CIFAR-10~\cite{cifar}. 
All our experiments are conducted by replacing original convolutions with the same channels of input and output feature maps.
And the new models are denoted as Comp-VGG16.
Since VGG-16 is originally designed for ImageNet, 
we adopt a little modification to the structure like~\cite{pytorch-cifar10} 
which is widely used in the literature to handle samples from CIFAR-10 in lower resolution.

\vspace{2pt}
\noindent{\textbf{Effect of Shuffle Block.}}
It's easily noticed that the output feature maps are obtained by integrating all recursive results in CompConv. 
As shown in Eq.~\eqref{eq::casFFT}, the divided part of feature maps, $\X_A$, identically maps few channels of primary features and transmits them to compose the output of the module.
To ensure $\X_A$ does not include fixed feature maps with the constant position of channels, we adopt a shuffle block~\cite{shufflenet} in $4$ groups after the concatenation. 
Tab.~\ref{tab:shuffle} compares the results between our CompConv and the one without shuffle block. 
The results imply that shuffle block can improve performance without adding any computations.
The shuffle operation in CompConv prevents propagating the same feature from shallow feature to the final output and enhance the feature representation ability.   
Thus we'll adopt the CompConv with shuffle block in the following experiments.

\setlength{\tabcolsep}{10pt}
\begin{table}[t]
    \vspace{0pt}
    \caption{\textbf{Effec of Shuffle Block.}
    We deploy CompConv on VGG-16 following Eq.~\eqref{eq::compute_p}.
    The proposed CompConv shuffle channels of feature maps after composition while the one without shuffle block keeps sequential concatenation. 
    }
    \label{tab:shuffle}
    \vspace{5pt}
    \centering\small
    \begin{tabular}{lccc}
    \toprule
        Model    & Params     & FLOPs  & Top-1($\%$)     \\ \midrule
        w.o Shuffle-Block & 3.3M      & 107M  & 92.8 \\
        \textbf{Ours}     & 3.3M      & 107M  & \textbf{93.8} \\
    \bottomrule
    \end{tabular}
    \vspace{0pt}
\end{table}

\setlength{\tabcolsep}{7pt}
\begin{table}[t]
    \vspace{0pt}
    \caption{\textbf{Effect of Identical Mapping}.
        We study the effectiveness of our \emph{fixed} identity mapping.
        The results of \emph{conv},\emph{random} and \emph{group} are shown for comparison with our \emph{fixed} identity mapping. 
        }
    \label{tab:copy}
    \vspace{5pt}
    \centering\small
    \begin{tabular}{llccc}
    \toprule
    Model  & Option   & Params     & FLOPs  & Top-1($\%$)   \\ \midrule
    \multirow{3}{*}{Comp-VGG16}
                & \emph{group}   & 3.3M     & 107M   & 93.7    \\
                & \emph{random}  & 3.3M     & 107M   & 92.7    \\  
                & \emph{conv}    & 3.6M     & 112M   & 93.8  \\ \midrule
    \textbf{Ours}   & \emph{fixed}   & 3.3M & 107M   & \textbf{93.8}    \\ 
    \bottomrule
    \end{tabular}
    \vspace{0pt}
\end{table}

\vspace{2pt}
\noindent{\textbf{Effect of Identical Mapping.}}
During recursive computation in CompConv, $\{\X_{A_{i}}, \quad i \in \{1...d\}\}$  is implemented by the identical mapping from the starting channels of input feature maps.
We conduct the following ablation to study the effect of the identity mapping options for $X_A$.
The basic operation to map starting channels from input feature map is denoted as \emph{fixed} in Tab.~\ref{tab:copy}.

We first give a comparison by generating these feature maps via applying $1\times 1$ convolution to the input feature map.
The \emph{conv} option brings more computations while the performance with $93.16\%$ is as same as the \emph{fixed} mapping option with $93.17\%$.

Except for the \emph{conv} option with larger parameters and FLOPs, 
we also give another two mapping options as group-selected and random-selected from the input feature without extra computation cost.
The \emph{group} option means dividing input channels by several groups then pick up specific feature maps from each group.
With the same parameters and FLOPs, feature sampling from group-division shows similar classification accuracy with \emph{fixed} option.
Intuitively, we think that such channels assignment drives the CompConv to seek principle feature maps in the input regardless of positions.
The \emph{random} option implies no fixed assignment but in a random mode with slightly lower performance than other mapping options with $0.4\%$ drop, 
Thus, we choose the simple way which incorporates the specific identical mapping from \emph{fixed} starting position of the input feature map.

\vspace{2pt}
\noindent{\textbf{Analysis on Recursion Depth $d$.}}
As shown in Eq.~\eqref{eq::complexity-compconv}, our CompConv has one hyper-parameter $d$ denotes recursion depth while larger $d$ brings greater computation compression. 
To explore the effect of different $d$, we firstly adopt a global setting of different $d$ by replacing all convolutions with CompConv. 
As shown in Tab.~\ref{tab::stage_n}, setting with global $d=1$ in CompConv can save half of parameters and FLOPs with $1\%$ loss of Top-1 accuracy.
When equipping CompConv with the global setting of $d=2$ or $d=3$, the whole network is compressed with higher efficiency but still maintains good performance.
However, for $d=4$ the performance presents a distinct drop compared with the other settings because of the severe squeeze for the primal channels based on Eq.~\eqref{eq::compute_p}. 
So we give the adaptive separation strategy shown in Sec. \ref{sec:method} on $d$, not exceeding 4 as written in Eq.~\eqref{eq::recommond_N}.

In addition, we substitute vanilla convolution at different stages to study the relation between $d$ and channels.
According to the four MaxPooling layers in VGG16, we divide the whole network into five stages.
We deploy CompConv only at the first stage with $64$ channels and last stage with $512$ channels to conduct this comparative study.
Comp-VGG16(1) indicates first stage replacement using CompConv by setting $d$ into different values from $\{1,2,3\}$ while other regular convolutions keep unchanged.
In this case, $d=1,2,3$ achieves nearly the same compression performance, 
while the Top-1 accuracy for $d=2,3$ is lower than $d=1$ with 0.7\%, which supports that the former stages of the network are less redundant.

When we deploy CompConv with $d=1,2,3$ only at the 5-th stage, it's noticed that parameters and FLOPs of the network decrease greatly, which still maintains considerable accuracy yet.
So we could use CompConv with a larger $d=3$ at the high-dimensional stage to achieve a better trade-off between computing expense and accuracy.
It reveals that there is more redundancy with larger channels.
Consequently, following Eq.~\eqref{eq::recommond_N}, the convolution with larger channels are equipped with larger recursive depth and vice versa.
Based on an adaptive selection of $d$ described in Sec.~\ref{subsec:separation-strategy}, our Comp-VGG16 demonstrates more significant performance than all with global $d$ settings as well as saving a large amount of parameters and FLOPs.
Thus we adopt the adaptive separation strategy for CompConv in the following experiments for effectiveness and
efficiency.

\setlength{\tabcolsep}{7pt}
\begin{table}[t]
    \caption{\textbf{Effect  of  Recursion  Depth $d$.}
    We conduct experiments on global settings with different $d$ in Comp-VGG16.
    SSAD denotes the Separation  Strategy  with  Adaptive  Depth.}
    \label{tab::global_n}
    \vspace{5pt}
    \centering\small
    \begin{tabular}{lcccc}
        \toprule
        Model                              & $d$  & Params     & FLOPs  & Top-1($\%$)   \\ \midrule
        VGG16                              & 0  & 14.7M     & 314M   &  94.1       \\ \midrule
        \multirow{4}{*}{Comp-VGG16}     & 1  & 7.4M      & 158M   &  93.0 \\
                                           & 2  & 4.3M      & 100M   &  92.7 \\
                                           & 3  & 2.9M      & 73M    &  92.6 \\
                                           & 4  & 2.2M      & 56M    &  92.0 \\ \midrule
        \textbf{Ours(w. SSAD)}                     & -  & 3.3M      & 107M   &  \textbf{93.8} \\            
        \bottomrule
    \end{tabular}
    \end{table}

\subsection{Image Classification} \label{main_results}

After studying the superiority of the proposed CompConv module for efficient feature learning, 
we then evaluate the well-designed CompConv architecture on image classification, including various datasets including CIFAR-10, CIFAR-100, and ImageNet. 

On CIFAR-10, we use the Comp-VGG16 model mentioned in Sec.~\ref{CIFAR-10-setting} to compare with other competitors. 
For ResNet on CIFAR-100 and ImageNet, we conduct experiments by replacing regular convolutions with CompConv except for the \textbf{Conv1} before the first MaxPooling. 
As a summary of former configurations, we adopt the recommended separation strategy as Eq.~\eqref{eq::recommond_N}. 
Unless specified otherwise, the training setting is as same as in Sec.~\ref{training-setting}.
Other than the model in full precision, we also integrate CompConv with a quantized network~\cite{rastegari2016xnor} to show the generality of CompConv.

\subsubsection{VGG on CIFAR-10}
We compare our Comp-VGG16 with some state-of-the-art models, including different types of model compression approaches. 
i.e., $l_1$ pruning~\cite{li2016pruning}, channel pruning~\cite{he2017channel}, HRank~\cite{HRANK}, SSS~\cite{sss}, SBP~\cite{he2017channel}.
In comparison, our Comp-VGG16 can achieve the best accuracy (only drops 0.3\% than the original one) with 3$\times$ acceleration and can save nearly about 5$\times$ storage.
Specifically, HRank~\cite{HRANK} can achieve the smallest number of parameters, while Comp-VGG16 saves 39M FLOPs with the 0.4\% improvement on Top-1 accuracy. 

\setlength{\tabcolsep}{7pt}
\begin{table}[t]
    \caption{\textbf{Effect of Recursion Depth $d$ on Different Stages.}
        CompConv with different recursive depth $d$ are adopted to the original VGG-16 at first stage and last stage of VGG-16 independently, named Comp-VGG16-1s and Comp-VGG16-5s.
        We use them to evaluate the effect of recursion depth $d$ on different stages.
      }
    \label{tab::stage_n}
    \vspace{5pt}
    \centering\small
    \begin{tabular}{lcccc}
        \toprule
        Model                           & $d$     & Params     & FLOPs  & Top-1($\%$)   \\ \midrule
        VGG16                           & 0       & 14.7M     & 314M   & 94.1    \\ \midrule
        \multirow{3}{*}{Comp-VGG16-1s}  & 1       & 14.7M     & 295M   & 93.8    \\
                                        & 2       & 14.7M     & 289M   & 93.1    \\
                                        &3       & 14.7M     & 287M   & 93.1    \\ \midrule
        \multirow{3}{*}{Comp-VGG16-5s}  & 1       &  8.2M     & 253M   & 93.8    \\
                                        & 2       &  5.5M     & 228M   & 93.7    \\
                                        & 3       &  4.3M     & 216M   & 93.6    \\
        \bottomrule
    \end{tabular}
\end{table}

\subsubsection{Binarized CompConv on CIFAR-10}
Except applying our CompConv to replace regular convolution in a full precision model, we also integrate CompConv with binary methods in XNORNet~\cite{rastegari2016xnor} 
to show the effectiveness of CompConv in the lightweight scenario.
We perform the same binarization process as shown in~\cite{rastegari2016xnor} for vanilla VGG-16, Ghost-VGG16 and our Comp-VGG16 and 
make an apple-to-apple comparison with these methods.
As shown in Tab.~\ref{Binary}, the binarized CompConv beat binarized Ghost both in parameters and FLOPs, with nearly 2$\times$ acceleration and storage saving.
Though the binarized model is lightweight enough, there still exists redundancy of computation cost as pointed by~\cite{xu2019main}, 
the computation efficiency of which can also be improved by replacing regular convolution with our CompConv.

\subsubsection{ResNet on CIFAR-100}
We also apply CompConv on ResNet to verify its capability of finer classification on CIFAR-100. 
The samples of CIFAR-100 are as same as CIFAR-10 but with finer labels. 
Here we mainly compare with another similar lightweight convolutional design Ghost Module.

We replace CompConv in both BasicBlock and BottleNeck respectively in ResNet18 and ResNet50 illustrated in Fig.~\ref{fig:compconv-res}.
According to results listed in Tab.~\ref{CIFAR-100}, our CompConv achieves better accuracy (nearly 1\%) but around $30\%$ FLOPs reduction than GhostModule. 
The results listed in Tab.~\ref{CIFAR-10} demonstrates that there is a tiny gap between the original model and substitute one with CompConv.
Intuitively, CIFAR-100 with finer labels is more difficult than CIFAR-10, which needs a larger model with more computation to maintain the feature representation.
In spite of this, we still obtain a good result with $40\%$ FLOPs and parameters of conventional ResNet at the cost of $3\%$ performance drop.

\begin{table}[t]
    \caption{Comparison of state-of-the-art methods for compressing VGG-16~\cite{vgg} on CIFAR-10~\cite{cifar} dataset.}
    \label{CIFAR-10}
    \vspace{5pt}
    \centering\small
    \begin{tabular}{lccc}
        \toprule
        Model           & Params             & FLOPs     & Top-1($\%$)\\ \midrule
        VGG16          ~\cite{vgg}           & 15M        & 314M      & 94.1     \\ \midrule
        L1-VGG16       ~\cite{li2016pruning} & 5.4M       & 206M      & 93.4     \\
        SSS-VGG16      ~\cite{sss}           & 3.9M       & 183M      & 93.0     \\
        Ghost-VGG16    ~\cite{ghostnet}      & 7.7M       & 158M      & 93.7     \\
        HRank-VGG16    ~\cite{HRANK}         & 2.5M       & 146M      & 93.4     \\
        SBP-VGG16      ~\cite{li2016pruning} &-          & 136M      & 92.5     \\
        \textbf{Comp-VGG16}  & \textbf{3.3M} & \textbf{107M}& \textbf{93.8} \\
        \bottomrule
    \end{tabular}
\end{table}

\begin{table}[t]
    \caption{Comparison with convolutional compressing design. We use another plug-and-play Ghost module~\cite{ghostnet} to make impartial comparison using the same binary method~\cite{rastegari2016xnor}.}
    \label{Binary}
    \vspace{5pt}
    \centering\small
    \begin{tabular}{lccc}
        \toprule
        Model  & Params & FLOPs     & Top-1($\%$) \\ \midrule
        Bin-VGG16         & 1.3M     & 22.4M      & 83.8     \\ \midrule
        Bin-Ghost-VGG16   & 0.9M     & 11.6M      & 79.1     \\
        \textbf{Bin-Comp-VGG16} & \textbf{0.3M}  & \textbf{6.6M} & \textbf{80.0} \\
        \bottomrule
    \end{tabular}
\end{table}

\subsubsection{ResNet on ImageNet}

\begin{figure}[t]
    \centering
    \includegraphics[width=1.0\columnwidth]{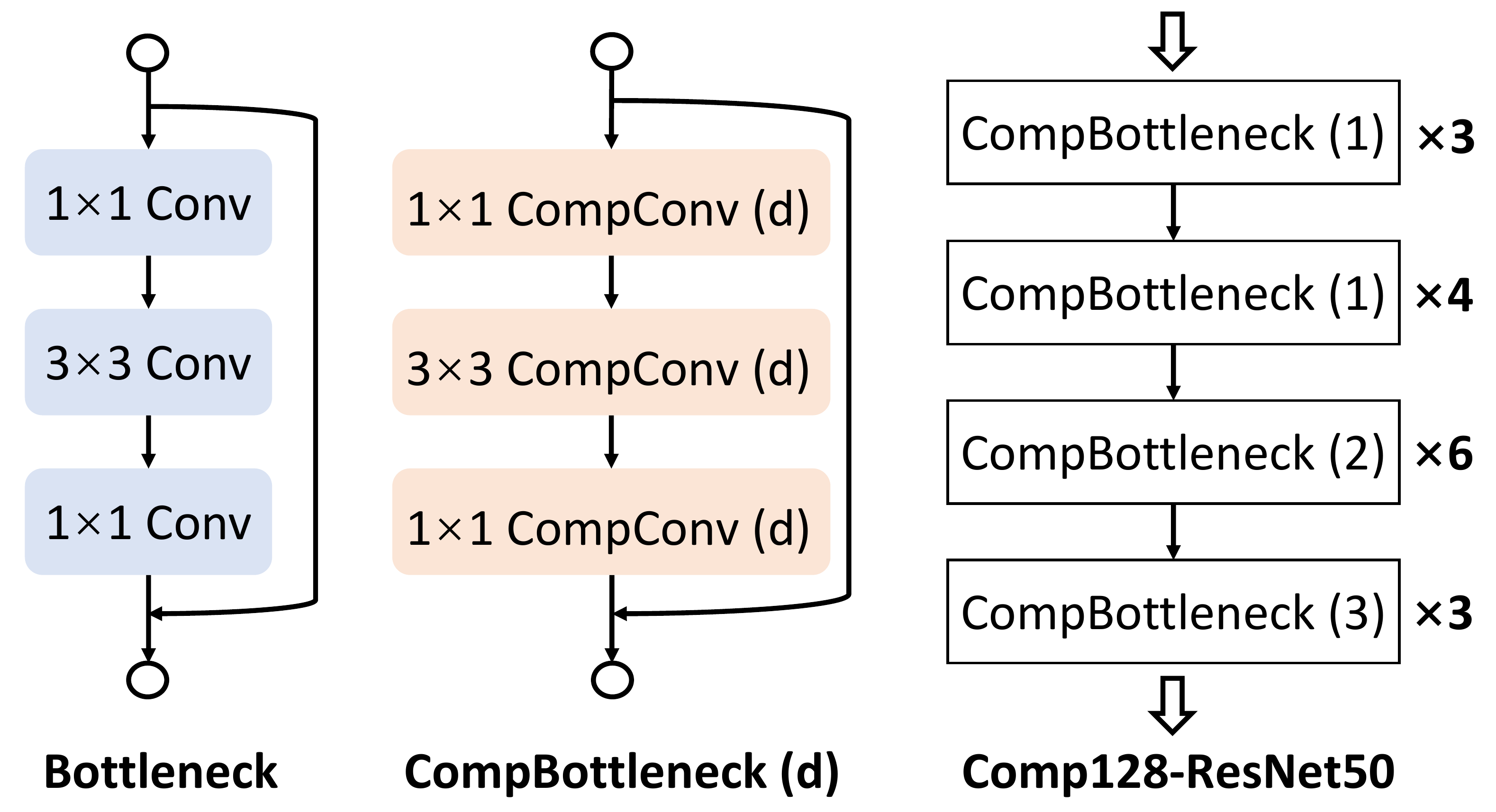}
    \caption{CompConv-ResNet50}
    \label{fig:compconv-res}
\end{figure}

Finally, We conduct experiments for ResNet-50 on the challenging ImageNet dataset, comparing with other state-of-the-art methods.
For this network structure, we replace all convolutions with CompConv as shown in Fig.~\ref{fig:compconv-res}, 
where we also take the example of separation strategy using Eq.~\eqref{eq::recommond_N} as CompConv128.

In order to give a comprehensive comparison on various FLOPs group, we apply three configurations by setting $c_0$ to 512, 256 and 128 in Eq.~\eqref{eq::recommond_N}, 
which are named as Comp512-ResNet50, Comp256-ResNet50, Comp128-ResNet50.

Compared with the recent state-of-the-art methods including Thinet\cite{luo2017thinet}, Versatile filters \cite{versatile} and Sparse structure selection \cite{sss} in larger FLOPs group,
our Comp512-ResNet50 beats them with $1.3\%$, $1.6\%$, $0.9\%$ promotion of accuracy.
Besides, our Comp512-ResNet50 keeps nearly the same performance as the original ResNet50 with 1.7$\times$ acceleration. 
When we further set $c_0$ to 256, our Comp256-ResNet50 has a 0.6\% accuracy drop while still obtains better results than HRank~\cite{HRANK} and GhostModule~\cite{ghostnet} with nearly the same FLOPs and fewer parameters.

For the lower FLOPs group, our Comp128-ResNet50 achieves the highest accuracy with the lowest FLOPs and parameters.
In contrast, compared methods including ABC-Prunner \cite{abcprunner} , GDP\cite{gdp}, GAL\cite{gal} with similar weights or FLOPs have much lower performance than ours.

All things considered, CompConv provides us with excellent compact convolutional design for CNN network.
\setlength{\tabcolsep}{12pt}
\begin{table}[t]
    \caption{Comparison of Ghost-ResNet~\cite{ghostnet} on CIFAR-100~\cite{cifar} dataset. We replace all convolutions in both basic block and bottleneck with Ghost module and CompConv within ResNet.}
    \label{CIFAR-100}
    \vspace{5pt}
    \centering\small
    \begin{tabular}{lccc}
        \toprule
        Model            & Params     & FLOPs     & Top-1($\%$) \\ \midrule
        Res18            & 11.2M      & 0.6G      & 76.4     \\ \midrule
        Ghost-Res18      & 5.7M       & 0.3G      & 72.3     \\
        \textbf{Comp-Res18}& \textbf{3.3M}  & \textbf{0.2G} & \textbf{73.1}     \\ \midrule[0.8pt]
        Res50            & 23.7M      & 1.3G      & 77.4     \\ \midrule
        Ghost-Res50      & 13.5M      & 0.7G      & 74.6     \\
        \textbf{Comp-Res50}   & \textbf{12.2M} & \textbf{0.5G} & \textbf{75.5}\\
        \bottomrule
    \end{tabular}
\end{table}

\begin{figure}[t]
    \centering
    \includegraphics[width=1.0\columnwidth]{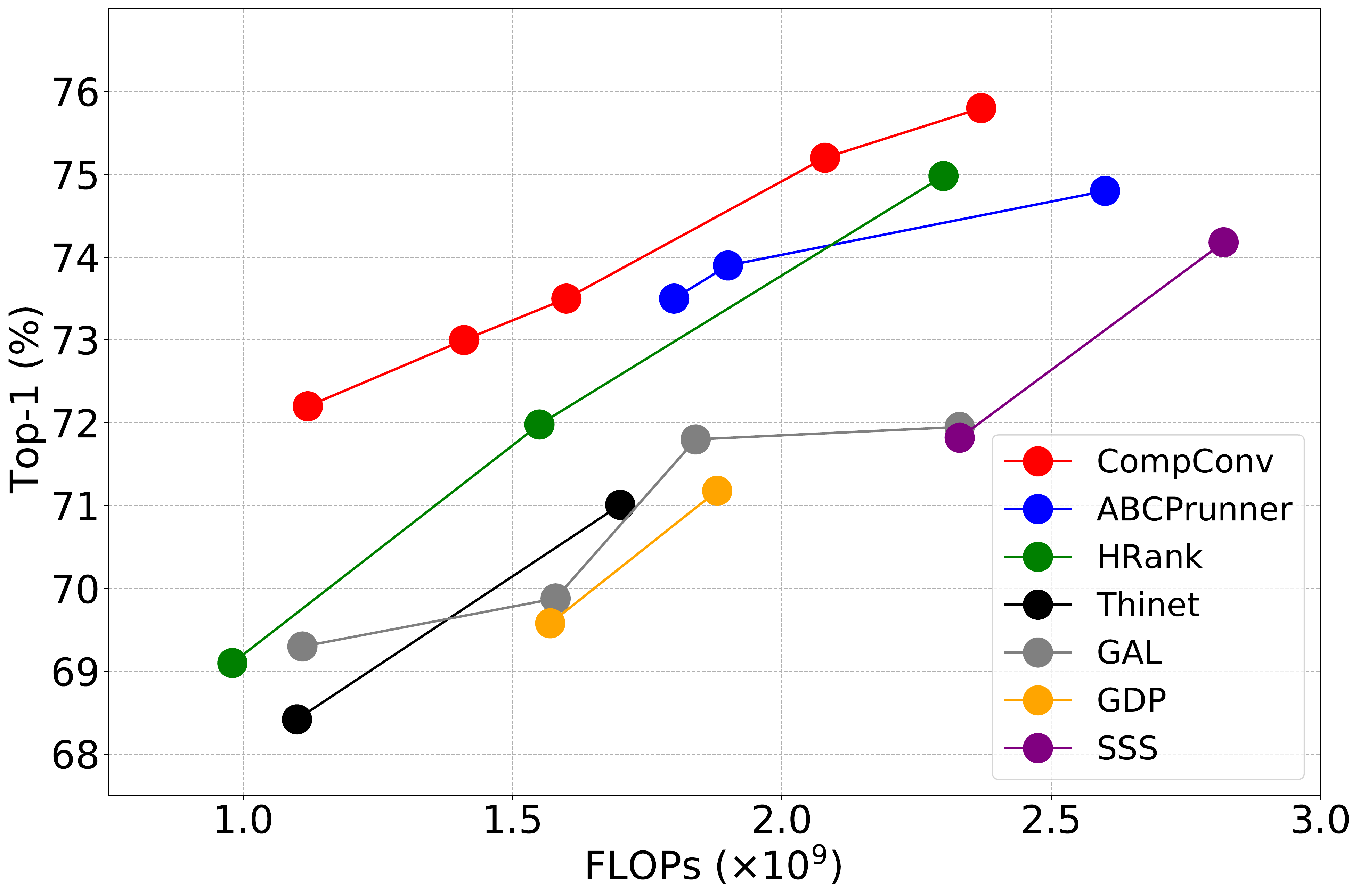}
    \caption{Top1-accuracy \emph{v.s.} FLOPs for ResNet-50 Model on ImageNet.}
    \label{fig:compare_all}
\end{figure}

\setlength{\tabcolsep}{20pt}
\begin{table*}[t]
    \caption{Comparison of state-of-the-art methods for compressing ResNet50 on ImageNet.}
    \label{imagenet}
    \vspace{5pt}
    \centering\small
    \begin{tabular}{lcccc}
        \toprule
        Model                             & Params  & FLOPs  & Top-1($\%$)& Top-5($\%$) \\ \midrule
        ResNet50          ~\cite{resnet}  & 25.6M   & 4.1G   & 76.1     & 92.9     \\ \midrule
        Versatile-ResNet50~\cite{versatile}  & 11.0M   & 3.0G   & 74.5     & 91.8     \\ 
        SSS-ResNet50-32~\cite{sss}           & 18.6M   & 2.8G   & 74.2     & 91.9     \\ 
        Thinet-ResNet50 ~\cite{luo2017thinet}& 16.9M   & 2.6G   & 72.1     & 90.3     \\ 
        \textbf{Comp512-ResNet50}            & \textbf{15.3M}  & \textbf{2.4G}  & \textbf{75.8}     & \textbf{92.8}     \\ \midrule
        HRank-ResNet50~\cite{HRANK}          & 16.2M   & 2.3G   & 75.0     & 92.3     \\ 
        NISP-ResNet50~\cite{nisp}            & 14.4M   & 2.3G   & -        & 90.8     \\ 
        Ghost-ResNet50-s2~\cite{ghostnet}    & 13.0M   & 2.2G   & 75.0     & 92.3     \\
        \textbf{Comp256-ResNet50}            & \textbf{13.7M}  & \textbf{2.1G}  & \textbf{75.2}     & \textbf{92.3}     \\ \midrule
        GDP-ResNet50-0.6 ~\cite{gdp}         & -       & 1.9G   & 71.2     & 90.7     \\
        GAL-ResNet50~\cite{gal}              & 19.3M   & 1.8G   & 71.8     & 90.8     \\ 
        ABC-ResNet50 ~\cite{abcprunner}      & 11.2M   & 1.8G   & 73.5     & 91.5    \\ 
        HRank-ResNet50~\cite{HRANK}          & 13.8M   & 1.6G   & 72.0     & 91.0     \\
        \textbf{Comp128-ResNet50}            & \textbf{8.7M}  & \textbf{1.6G}  & \textbf{73.7}     & \textbf{91.2}              \\
        \bottomrule
    \end{tabular}
\end{table*}
Since we've talked thoroughly about the influence of recursion depth $d$ when using CompConv, 
we adopt recommended separation strategy based on Eq.~\eqref{eq::recommond_N} with $c_0=512,256,128,64,32$ to get more results in Fig.~\ref{fig:compare_all}. 
This solution gives a dynamic comparison by setting CompConv-ResNet50 with different computation-consuming levels. 
For comparison, we select several state-of-the-art methods to draw the FLOPs v.s. Top-1 accuracy curve. 
From an overall view of Fig.~\ref{fig:compare_all}, the curve of our CompConv are above all other methods including SSS\cite{sss}, Thinet\cite{luo2017thinet}, GDP\cite{gdp} and GAL\cite{gal}, ABC-Prunner \cite{abcprunner}, HRANK \cite{HRANK}.
It shows that our CompConv gains higher accuracy with the same computation cost.
As for the same performance, the model with our CompConv performs more compact than others.

\subsection{Object Detection} \label{detection}
In addition to results on classification benchmarks,
 we also evaluate our CompConv in object detection on MS COCO dataset~\cite{coco}. 
 We deploy our CompConv to the state-of-the-art detector Faster-RCNN~\cite{faterrcnn} following the train and inference setting in Sec.~\ref{subsec:separation-strategy}.
As shown in Tab.~\ref{tab:detection} where the FLOPs are calculated using 800 $\times$ 1333 images following MMDetection~\cite{mmdetection}, 
our Comp512-ResNet50 achieve nearly the same performance with lower computational costs compared with vanilla ResNet50. 
The smaller model Comp128-ResNet50 can save 47G FLOPs and 11.43M parameters cost of a slight drop on mAP.

\setlength{\tabcolsep}{10pt}
\begin{table}[t]
    \caption{Results on COCO with  Faster-RCNN detector.}
    \label{tab:detection}
    \vspace{5pt}
    \centering\small
    \begin{tabular}{lccc}
    \toprule
        Model                &  Params & FLOPs & mAP \\ \midrule
        ResNet50             & 41.5M & 207.1G & 37.4 \\ \midrule
        Comp512-ResNet50    & 31.5M & 176.3G & 37.2 \\
        Comp128-ResNet50    & 30.1M & 160.8G & 36.6 \\ \bottomrule
    \end{tabular}
\end{table}

\section{Discussion and Conclusion}
To reduce the computational costs of CNN, this paper presents a novel design denoted as CompConv from perspective of the basic convolution unit. It utilizes divide-and-conquer strategy to simplify transformation of feature maps. The proposed CompConv is a plug-and-play module for replacing vanilla convolution without additional restrictions. Comprehensive experiments conducted on various visual benchmark models illustrate that our CompConv greatly save FLOPs and parameters while still keeping a considerable accuracy performance with good transferability. It's an efficient convolutional design to make CNNs lighter to deploy on resources-limited ends.

One important future work direction is to design a more adaptive and comprehensive solution when applying CompConv to wider network. The recursion depth $d$ accounts a lot in saving computation and enhancing performance. Maybe we can use AutoML technology to always find out the most suitable settings when applying to different networks. In that way, CompConv would play a stable role in any convolutional network structures more easily.
Besides, how to deploy the CompConv in a smaller model \ie the pruned model or the quantized model to further compress the network remains an open problem.

{\small
\bibliographystyle{ieee_fullname}
\bibliography{ref}

\begin{thebibliography}{10}\itemsep=-1pt

\bibitem{bender2018understanding}
Gabriel Bender, Pieter-Jan Kindermans, Barret Zoph, Vijay Vasudevan, and Quoc
  Le.
\newblock Understanding and simplifying one-shot architecture search.
\newblock In {\em ICML}, 2018.

\bibitem{cai2019once}
Han Cai, Chuang Gan, Tianzhe Wang, Zhekai Zhang, and Song Han.
\newblock Once-for-all: Train one network and specialize it for efficient
  deployment.
\newblock {\em arXiv preprint arXiv:1908.09791}, 2019.

\bibitem{mmdetection}
Kai Chen, Jiaqi Wang, Jiangmiao Pang, Yuhang Cao, Yu Xiong, Xiaoxiao Li,
  Shuyang Sun, Wansen Feng, Ziwei Liu, Jiarui Xu, Zheng Zhang, Dazhi Cheng,
  Chenchen Zhu, Tianheng Cheng, Qijie Zhao, Buyu Li, Xin Lu, Rui Zhu, Yue Wu,
  Jifeng Dai, Jingdong Wang, Jianping Shi, Wanli Ouyang, Chen~Change Loy, and
  Dahua Lin.
\newblock {MMDetection}: Open mmlab detection toolbox and benchmark.
\newblock {\em arXiv preprint arXiv:1906.07155}, 2019.

\bibitem{courbariaux2015binaryconnect}
Matthieu Courbariaux, Yoshua Bengio, and Jean-Pierre David.
\newblock Binaryconnect: Training deep neural networks with binary weights
  during propagations.
\newblock In {\em Adv. Neural Inform. Process. Syst.}, 2015.

\bibitem{imagenet}
Jia Deng, Wei Dong, Richard Socher, Li-Jia Li, Kai Li, and Li Fei-Fei.
\newblock Imagenet: A large-scale hierarchical image database.
\newblock In {\em IEEE Conf. Comput. Vis. Pattern Recog.}, 2009.

\bibitem{dong2017learning}
Xin Dong, Shangyu Chen, and Sinno Pan.
\newblock Learning to prune deep neural networks via layer-wise optimal brain
  surgeon.
\newblock In {\em Adv. Neural Inform. Process. Syst.}, 2017.

\bibitem{limitednumerical}
Suyog Gupta, Ankur Agrawal, Kailash Gopalakrishnan, and Pritish Narayanan.
\newblock Deep learning with limited numerical precision.
\newblock In {\em ICML}, 2015.

\bibitem{co-teaching}
Bo Han, Quanming Yao, Xingrui Yu, Gang Niu, Miao Xu, Weihua Hu, Ivor Tsang, and
  Masashi Sugiyama.
\newblock Co-teaching: Robust training of deep neural networks with extremely
  noisy labels.
\newblock In {\em Adv. Neural Inform. Process. Syst.}, 2018.

\bibitem{ghostnet}
Kai Han, Yunhe Wang, Qi Tian, Jianyuan Guo, Chunjing Xu, and Chang Xu.
\newblock Ghostnet: More features from cheap operations.
\newblock In {\em IEEE Conf. Comput. Vis. Pattern Recog.}, 2020.

\bibitem{han16pruning}
Song Han, Huizi Mao, and William~J Dally.
\newblock Deep compression: Compressing deep neural networks with pruning,
  trained quantization and huffman coding.
\newblock {\em Int. Conf. Learn. Represent.}, 2015.

\bibitem{resnet}
Kaiming He, Xiangyu Zhang, Shaoqing Ren, and Jian Sun.
\newblock Deep residual learning for image recognition.
\newblock In {\em IEEE Conf. Comput. Vis. Pattern Recog.}, 2016.

\bibitem{he2017channel}
Yihui He, Xiangyu Zhang, and Jian Sun.
\newblock Channel pruning for accelerating very deep neural networks.
\newblock In {\em Int. Conf. Comput. Vis.}, 2017.

\bibitem{distilling}
Geoffrey Hinton, Oriol Vinyals, and Jeff Dean.
\newblock Distilling the knowledge in a neural network.
\newblock {\em arXiv preprint arXiv:1503.02531}, 2015.

\bibitem{mobilenetv3}
Andrew Howard, Mark Sandler, Grace Chu, Liang-Chieh Chen, Bo Chen, Mingxing
  Tan, Weijun Wang, Yukun Zhu, Ruoming Pang, Vijay Vasudevan, et~al.
\newblock Searching for mobilenetv3.
\newblock In {\em Int. Conf. Comput. Vis.}, 2019.

\bibitem{mobilenets}
Andrew~G Howard, Menglong Zhu, Bo Chen, Dmitry Kalenichenko, Weijun Wang,
  Tobias Weyand, Marco Andreetto, and Hartwig Adam.
\newblock Mobilenets: Efficient convolutional neural networks for mobile vision
  applications.
\newblock {\em arXiv preprint arXiv:1704.04861}, 2017.

\bibitem{huang2017like}
Zehao Huang and Naiyan Wang.
\newblock Like what you like: Knowledge distill via neuron selectivity
  transfer.
\newblock {\em arXiv preprint arXiv:1707.01219}, 2017.

\bibitem{sss}
Zehao Huang and Naiyan Wang.
\newblock Data-driven sparse structure selection for deep neural networks.
\newblock In {\em Eur. Conf. Comput. Vis.}, 2018.

\bibitem{squeezenet}
Forrest~N Iandola, Song Han, Matthew~W Moskewicz, Khalid Ashraf, William~J
  Dally, and Kurt Keutzer.
\newblock Squeezenet: Alexnet-level accuracy with 50x fewer parameters and< 0.5
  mb model size.
\newblock {\em arXiv preprint arXiv:1602.07360}, 2016.

\bibitem{jacob2018quantization}
Benoit Jacob, Skirmantas Kligys, Bo Chen, Menglong Zhu, Matthew Tang, Andrew
  Howard, Hartwig Adam, and Dmitry Kalenichenko.
\newblock Quantization and training of neural networks for efficient
  integer-arithmetic-only inference.
\newblock In {\em IEEE Conf. Comput. Vis. Pattern Recog.}, 2018.

\bibitem{cifar}
Alex Krizhevsky et~al.
\newblock Learning multiple layers of features from tiny images.
\newblock 2009.

\bibitem{alexnet}
Alex Krizhevsky, Ilya Sutskever, and Geoffrey~E Hinton.
\newblock Imagenet classification with deep convolutional neural networks.
\newblock {\em Communications of the ACM}, 2012.

\bibitem{li2016pruning}
Hao Li, Asim Kadav, Igor Durdanovic, Hanan Samet, and Hans~Peter Graf.
\newblock Pruning filters for efficient convnets.
\newblock {\em Int. Conf. Learn. Represent.}, 2017.

\bibitem{HRANK}
Mingbao Lin, Rongrong Ji, Yan Wang, Yichen Zhang, Baochang Zhang, Yonghong
  Tian, and Ling Shao.
\newblock Hrank: Filter pruning using high-rank feature map.
\newblock In {\em IEEE Conf. Comput. Vis. Pattern Recog.}, 2020.

\bibitem{abcprunner}
Mingbao Lin, Rongrong Ji, Yuxin Zhang, Baochang Zhang, Yongjian Wu, and
  Yonghong Tian.
\newblock Channel pruning via automatic structure search.
\newblock {\em arXiv preprint arXiv:2001.08565}, 2020.

\bibitem{gdp}
Shaohui Lin, Rongrong Ji, Yuchao Li, Yongjian Wu, Feiyue Huang, and Baochang
  Zhang.
\newblock Accelerating convolutional networks via global \& dynamic filter
  pruning.
\newblock In {\em IJCAI}, 2018.

\bibitem{gal}
Shaohui Lin, Rongrong Ji, Chenqian Yan, Baochang Zhang, Liujuan Cao, Qixiang
  Ye, Feiyue Huang, and David Doermann.
\newblock Towards optimal structured cnn pruning via generative adversarial
  learning.
\newblock In {\em IEEE Conf. Comput. Vis. Pattern Recog.}, 2019.

\bibitem{coco}
Tsung-Yi Lin, Michael Maire, Serge Belongie, James Hays, Pietro Perona, Deva
  Ramanan, Piotr Doll{\'a}r, and C~Lawrence Zitnick.
\newblock Microsoft coco: Common objects in context.
\newblock In {\em Eur. Conf. Comput. Vis.}, 2014.

\bibitem{liu2018progressive}
Chenxi Liu, Barret Zoph, Maxim Neumann, Jonathon Shlens, Wei Hua, Li-Jia Li, Li
  Fei-Fei, Alan Yuille, Jonathan Huang, and Kevin Murphy.
\newblock Progressive neural architecture search.
\newblock In {\em Eur. Conf. Comput. Vis.}, 2018.

\bibitem{pytorch-cifar10}
Kuang Liu.
\newblock Pyotrch cifar10.
\newblock \url{https://github.com/kuangliu/pytorch-cifar.git}, 2019.

\bibitem{Liu2017learning}
Zhuang Liu, Jianguo Li, Zhiqiang Shen, Gao Huang, Shoumeng Yan, and Changshui
  Zhang.
\newblock Learning efficient convolutional networks through network slimming.
\newblock In {\em Int. Conf. Comput. Vis.}, 2017.

\bibitem{luo2017thinet}
Jian-Hao Luo, Jianxin Wu, and Weiyao Lin.
\newblock Thinet: A filter level pruning method for deep neural network
  compression.
\newblock In {\em Int. Conf. Comput. Vis.}, 2017.

\bibitem{shufflenetv2}
Ningning Ma, Xiangyu Zhang, Hai-Tao Zheng, and Jian Sun.
\newblock Shufflenet v2: Practical guidelines for efficient cnn architecture
  design.
\newblock In {\em Eur. Conf. Comput. Vis.}, 2018.

\bibitem{rastegari2016xnor}
Mohammad Rastegari, Vicente Ordonez, Joseph Redmon, and Ali Farhadi.
\newblock Xnor-net: Imagenet classification using binary convolutional neural
  networks.
\newblock In {\em Eur. Conf. Comput. Vis.}, 2016.

\bibitem{faterrcnn}
Shaoqing Ren, Kaiming He, Ross Girshick, and Jian Sun.
\newblock Faster r-cnn: Towards real-time object detection with region proposal
  networks.
\newblock In {\em Adv. Neural Inform. Process. Syst.}, 2015.

\bibitem{mobilenetv2}
Mark Sandler, Andrew Howard, Menglong Zhu, Andrey Zhmoginov, and Liang-Chieh
  Chen.
\newblock Mobilenetv2: Inverted residuals and linear bottlenecks.
\newblock In {\em IEEE Conf. Comput. Vis. Pattern Recog.}, 2018.

\bibitem{vgg}
Karen Simonyan and Andrew Zisserman.
\newblock Very deep convolutional networks for large-scale image recognition.
\newblock {\em arXiv preprint arXiv:1409.1556}, 2014.

\bibitem{versatile}
Yunhe Wang, Chang Xu, XU Chunjing, Chao Xu, and Dacheng Tao.
\newblock Learning versatile filters for efficient convolutional neural
  networks.
\newblock In {\em Adv. Neural Inform. Process. Syst.}, 2018.

\bibitem{xu2019main}
Yinghao Xu, Xin Dong, Yudian Li, and Hao Su.
\newblock A main/subsidiary network framework for simplifying binary neural
  networks.
\newblock In {\em IEEE Conf. Comput. Vis. Pattern Recog.}, 2019.

\bibitem{yim2017gift}
Junho Yim, Donggyu Joo, Jihoon Bae, and Junmo Kim.
\newblock A gift from knowledge distillation: Fast optimization, network
  minimization and transfer learning.
\newblock In {\em IEEE Conf. Comput. Vis. Pattern Recog.}, 2017.

\bibitem{nisp}
Ruichi Yu, Ang Li, Chun-Fu Chen, Jui-Hsin Lai, Vlad~I Morariu, Xintong Han,
  Mingfei Gao, Ching-Yung Lin, and Larry~S Davis.
\newblock Nisp: Pruning networks using neuron importance score propagation.
\newblock In {\em IEEE Conf. Comput. Vis. Pattern Recog.}, 2018.

\bibitem{zagoruyko2016paying}
Sergey Zagoruyko and Nikos Komodakis.
\newblock Paying more attention to attention: Improving the performance of
  convolutional neural networks via attention transfer.
\newblock {\em arXiv preprint arXiv:1612.03928}, 2016.

\bibitem{shufflenet}
Xiangyu Zhang, Xinyu Zhou, Mengxiao Lin, and Jian Sun.
\newblock Shufflenet: An extremely efficient convolutional neural network for
  mobile devices.
\newblock In {\em IEEE Conf. Comput. Vis. Pattern Recog.}, 2018.

\bibitem{zhong2018practical}
Zhao Zhong, Junjie Yan, Wei Wu, Jing Shao, and Cheng-Lin Liu.
\newblock Practical block-wise neural network architecture generation.
\newblock In {\em IEEE Conf. Comput. Vis. Pattern Recog.}, 2018.

\bibitem{zhou2017incremental}
Aojun Zhou, Anbang Yao, Yiwen Guo, Lin Xu, and Yurong Chen.
\newblock Incremental network quantization: Towards lossless cnns with
  low-precision weights.
\newblock {\em arXiv preprint arXiv:1702.03044}, 2017.

\bibitem{zoph2016neural}
Barret Zoph and Quoc~V Le.
\newblock Neural architecture search with reinforcement learning.
\newblock {\em arXiv preprint arXiv:1611.01578}, 2016.

\bibitem{zoph2018learning}
Barret Zoph, Vijay Vasudevan, Jonathon Shlens, and Quoc~V Le.
\newblock Learning transferable architectures for scalable image recognition.
\newblock In {\em IEEE Conf. Comput. Vis. Pattern Recog.}, 2018.

\end{thebibliography}
}

\end{document}